\documentclass{article}
\usepackage{spconf,amsmath,graphicx}
\usepackage{booktabs} 
\usepackage{diagbox} 
\usepackage{multirow}
\usepackage{subfig}
\usepackage{url}
\usepackage{hyperref}
\usepackage{amssymb}
\usepackage{balance}
\usepackage{pifont}
\newcommand{\cmark}{\ding{51}}
\newcommand{\xmark}{\ding{55}}
\title{Context-aware Pedestrian Trajectory Prediction with\\Multimodal Transformer}%
\name{Haleh Damirchi\qquad Michael Greenspan\qquad Ali Etemad} \address{Dept. ECE \& Ingenuity Labs Research Institute, Queen's University, Kingston, Canada} 

\begin{document}
%\ninept
%
\maketitle
\begin{abstract}
% In this paper, 
We propose a novel solution for predicting future trajectories of pedestrians. Our method uses a multimodal encoder-decoder transformer architecture, which takes as input both pedestrian locations and ego-vehicle speeds. Notably, our decoder predicts the entire future trajectory in a single-pass and does not perform one-step-ahead prediction, which makes the method effective for embedded edge deployment. We perform detailed experiments and evaluate our method on two popular datasets, PIE and JAAD. Quantitative results demonstrate the superiority of our proposed model over the current state-of-the-art, which consistently achieves the lowest error for 3 time horizons of 0.5, 1.0 and 1.5 seconds. Moreover, the proposed method is significantly faster than the state-of-the-art for the two datasets of PIE and JAAD. Lastly, ablation experiments demonstrate the impact of the key multimodal configuration of our method.
\end{abstract}
\section{Introduction}
\label{sec:intro}

Pedestrian trajectory prediction enables self-driving cars to predict the future motion of pedestrians, and has the potential to enhance the safety of both pedestrians and drivers by preventing dangerous scenarios. The performance of this task depends on numerous environmental factors, including the presence of neighboring pedestrians and vehicles, the speed of the vehicle, and environmental conditions.

Various recent solutions have been proposed to address pedestrian trajectory prediction. Being a time-series forecasting problem in nature, Long Short-Term Memory (LSTM) ~\cite{hochreiter1997long} and Gated Recurrent Unit (GRU)~\cite{chung2014empirical} networks have been widely used for this task~\cite{bayesianlstm, sgnet}. Recently, transformers have shown outstanding results on problems involving time-series representation learning, for instance in natural language~\cite{attention,bert}, videos~\cite{vivit, survey}, and physiological signals~\cite{Behinaein2021}. However, transformers are large data-hungry networks that often suffer from overfitting, and require delicate parameter tuning.

To predict pedestrian trajectories reliably, it is beneficial to leverage contextual information from the environment where possible~\cite{pie, bayesianlstm,crowdcounting}. Contextual information for pedestrian trajectory prediction from a car's point of view can be visual (e.g. images) or non-visual (e.g. vehicle speed). For instance, it has been shown in~\cite{bayesianlstm} that non-visual cues can yield significant improvements when combined with the trajectory information. Yet, despite the strong ability of transformers in learning contextual information as shown in~\cite{khaleghi2022learning}, they have only rarely been used to tackle this problem in the field of pedestrian trajectory prediction~\cite{tranforemrtraj}.

In this paper, we propose a novel transformer-based solution for pedestrian trajectory prediction with a multimodal encoder to leverage the information from both the observed pedestrian  trajectory and ego-vehicle speed. 
% In contrast to many existing time-series prediction solutions [??, ??] that generate the new data one future time step at a time (using iterative strategies that employ past prediction values as inputs), our solution uses a single-pass approach to generating the entire set of future time steps comprising the predicted trajectory at once. 
Many existing time-series prediction solutions~\cite{hochreiter1997long,chung2014empirical,attention} generate the new data one future time step at a time, using iterative strategies that employ past prediction values as inputs. In contrast, our solution uses a single-pass approach to generating the entire set of future time steps comprising the predicted trajectory at once. As a result, our method provides a lightweight and robust solution suitable for edge deployment in mobile and vehicular technologies. Detailed experiments and comparison to the state-of-the-art~\cite{sgnet} show that our proposed architecture can predict future trajectories more effectively with fewer hyperparameters and a considerably reduced inference time. 

Our contributions can be summarized as follows.
(\textbf{1}) We propose a single-pass transformer-based encoder-decoder architecture to predict the trajectory of pedestrians from the car's point of view. We treat the vehicle speed as contextual information and use it as an auxiliary input to our multimodal solution to improve the performance of predicted pedestrian trajectories. By jointly encoding and leveraging observed trajectory and vehicle ego-speed values, our network can accommodate for changes in the perceived trajectory of pedestrians caused by the variation in the vehicle's speed.
(\textbf{2}) Detailed experiments demonstrate that our method outperforms the state-of-the-art on two popular benchmarks, PIE~\cite{pie} and JAAD~\cite{jaad}. Moreover, our proposed architecture requires significantly less inference time. 
% (\textbf{3}) Upon acceptance, we will make our code public: \href{https://queensuca-my.sharepoint.com/:u:/g/personal/20hd25_queensu_ca/EQKrHL6dfRBBvX6lPqUNxwwBkDQPfHlre60RikuEtpFP6g?e=sE2noN}{link}.
(\textbf{3}) We make the code publicly available at: https://github.com/thisishale/Context-Aware.

\begin{figure*}[!t]
  \centering
  \includegraphics[width=1.73\columnwidth]
  % {images/transformer_sp_traj_loop_v7.pdf}
  {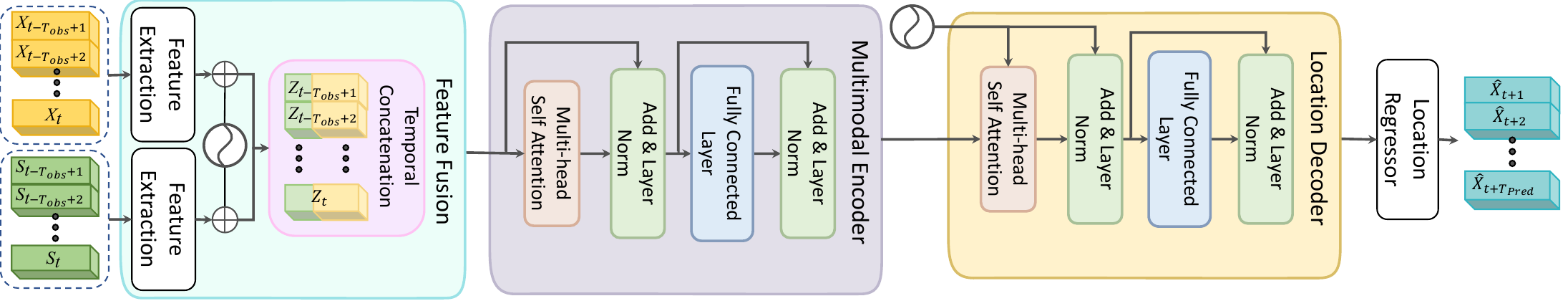}
  \caption{The proposed multimodal encoder-decoder transformer architecture. $\{X_i\}_{t-T_{obs}+1}^{t}$, $\{S_i\}_{t-T_{obs}+1}^{t}$, $\{Z_i\}_{t-T_{obs}+1}^{t}$ and $\{\widehat{X}_j\}_{t+1}^{t+T_{pred}}$ denote the observed pedestrian location, the observed ego-vehicle speeds, the concatenated features of both observed location and ego-vehicle speed and the predicted pedestrian location, respectively.}
  \label{fig:proposed}
\end{figure*}

\section{Related Work}
\label{sec:format}

Literature in the area can be categorized into two groups, \textit{bird's-eye view} and \textit{first-person view}, where our method falls in the latter category. Following we describe the related work in each group.

\noindent\textbf{Bird's-eye view trajectory prediction.}
This category of prediction focuses on forecasting pedestrian trajectories from a bird's-eye view. In one of the early solutions to use LSTM, each pedestrian trajectory was predicted by an LSTM network~\cite{sociallstm}. The LSTMs of neighboring trajectories are all connected to each other through pooling of hidden states. Later in ~\cite{mangalam2021goals}, scene segmentation maps along with past trajectory heatmaps were used as inputs to an encoder-decoder Convolutional Neural Network (CNN), to predict trajectory endpoint heatmaps, followed by complete trajectories. Finally, an encoder-decoder transformer network was recently used in~\cite{tranforemrtraj} to predict the future trajectory of pedestrians.

% Y-Net~\cite{mangalam2021goals} uses scene segmentation maps to predict heat maps of probable goal locations for future trajectory prediction. 

\noindent\textbf{First-person view trajectory prediction.}
Due to the parallax caused by the ego-vehicle speed and increased impact of the perspective effect, predicting pedestrian trajectories from a car's point of view is more challenging than the bird's-eye view approach. To tackle this problem, recurrent neural networks (RNNs) have been a key component in the majority of solutions in the area. For instance, an LSTM encoder-decoder network in a method called Bayesian LSTM predicted future vehicle ego-motion and combined it with the encoded observed trajectory to predict future pedestrian trajectories~\cite{bayesianlstm}. In this work, epistemic uncertainty was also estimated as an indication of the model's confidence. In PIE-Traj~\cite{pie}, visual local context (images) were used to estimate pedestrian intention (e.g., to cross the street) which was further employed together with decoded future vehicle speeds to estimate the future trajectory of the pedestrian. Next, in SGNet~\cite{sgnet}, an attention network was used to generate a sequence of intermediate goals which were used as inputs to a GRU network, in order to predict pedestrian trajectories. Another GRU encoder-decoder architecture has was proposed in BiTraP~\cite{bitrap}, which leverages a bi-directional decoder. Finally, a model consisting of 3 encoders, one for spatio-temporal features from optical flow, one for observed locations, and another for vehicle ego-motion information was proposed in FOL-X~\cite{folx}.% Location and spatio-temporal encoders were used in conjunction with decoded vehicle ego-motion information in order to predict the future pedestrian locations.

\begin{table*}[!t]
\centering
\footnotesize
\caption{Evaluation results for PIE and JAAD datasets. SGNet* denotes that the SGNet model is \textit{retrained}.}
\resizebox{1.4\columnwidth}{!}{
\begin{tabular}{ccccccccccc}
% \toprule[0.3pt]\midrule[0.3pt]
\hline
   & \multicolumn{5}{c}{\textbf{PIE}} & 
\multicolumn{5}{c}{\textbf{JAAD}}  \\
\cmidrule(r){2-6} \cmidrule(r){7-11}
  Method & \multicolumn{3}{c}{MSE} &  CMSE & CFMSE &
\multicolumn{3}{c}{MSE} &  CMSE & CFMSE  \\
\cmidrule(r){2-4} \cmidrule(r){7-9}
  & 0.5s & 1s & 1.5s & 1.5s & 1.5s & 0.5s & 1s & 1.5s & 1.5s & 1.5s\\
\midrule
% \hline
Bayesian LSTM~\cite{bayesianlstm} & 101 & 296 & 855 & 822 & 3259 & 159 & 539 & 1535 & 1447 & 5615   \\ 
FOL-X~\cite{folx} & 147 & 484 & 1374 & 546 & 2303 & 147 & 484 & 1374 & 1290 & 4924 \\ 
PIE-Traj~\cite{pie} & 58 & 200 & 636 & 596 & 2477 & 110 & 399 & 1248 & 1183 & 4780\\ 
BiTraP~\cite{bitrap} & 41 & 161 & 511 & 481 & 1949 & 93 & 378 & 1206 & 1105 & 4565\\ 
SGNet~\cite{sgnet} & \underline{34} & \underline{133} & \underline{442} & \underline{413} & \underline{1761} & \underline{82} & \underline{328} & \underline{1049} & \underline{996} & \underline{4076}\\ 
SGNet*~\cite{sgnet} & 36 & 139 & 459 & 429 & 1841 & \underline{82} & \underline{328} & 1051 & 1001 & 4098\\ 
Transformer~\cite{attention} & 52 & 211 & 661 & 624 & 2517 & 100 & 455 & 1672 & 1609 & 7182\\ 
% CPTNet$_{ES}$ 
% CPTNet(early summ.) & \textbf{34} & \textbf{128} & \textbf{397} & \textbf{1529} & \textbf{372} & 92 & 369 & 1143 & 4356 & 1091 \\
% CPTNet$_{CA1}$ 
% CPTNet (cross att.) & 35 & \textbf{129} & \textbf{402} & \textbf{377} & \textbf{1544} & 85 & 340 & 1063 & 1014 & 4034 \\
% CPTNet(cross att. 2) & - & - & - & - & - & - & - & - & - & - \\
% CPTNet$_{EC}$ 
% CPTNet (early concat.) & \textbf{34} & \textbf{130} & \textbf{408} & \textbf{383} & \textbf{1597} & \textbf{80} & \textbf{316} & \textbf{1008} & \textbf{961} & \textbf{3867}\\ 
Proposed & \textbf{33} & \textbf{127} & \textbf{398} & \textbf{372} & \textbf{1519} & \textbf{78} & \textbf{324} & \textbf{1020} & \textbf{974} & \textbf{3937}\\ 
% CPTNet-45 & \underline{34} & \underline{128} & \underline{408} & \underline{383} & \underline{1597} & \underline{81} & \textbf{322} & \textbf{1001} & \textbf{952} & \textbf{3795}\\ 
% \bottomrule
\hline
\label{table:eval_res}
\end{tabular}}
\end{table*}

 \begin{table*}[!t]
\centering
\footnotesize
\caption{Ablation studies on PIE and JAAD datasets.}
\resizebox{1.4\columnwidth}{!}{
\begin{tabular}{ccccccccccccc}
% \toprule[1pt]\midrule[0.3pt]
\hline
   &   & \multicolumn{5}{c}{\textbf{PIE}} & 
\multicolumn{5}{c}{\textbf{JAAD}}  \\
\cmidrule(r){3-7} \cmidrule(r){8-12}
 \multicolumn{2}{c}{Method}& \multicolumn{3}{c}{MSE} &  CMSE & CFMSE &
\multicolumn{3}{c}{MSE} &  CMSE & CFMSE  \\
 \cmidrule(r){1-2} \cmidrule(r){3-5} \cmidrule(r){8-10}
 Speed & Location & 0.5s & 1s & 1.5s & 1.5s & 1.5s & 0.5s & 1s & 1.5s & 1.5s & 1.5s\\
\midrule
\cmark &  \cmark  & \textbf{33} & \textbf{127} & \textbf{398} & \textbf{372} & \textbf{1519} & \underline{78} & \underline{324} & \textbf{1020} & \textbf{974} & \textbf{3937}\\
\xmark & \cmark  & \textbf{33} & \underline{133} & \underline{434} & \underline{407} & \underline{1712} & \textbf{76} & \textbf{319} & \underline{1039} & \underline{992} & \underline{4068}   \\ 
\cmark & \xmark  & 447 & 1769 & 4317 & 4238 & 13731 & 815 & 3225 & 7720 & 7617 & 24065  \\ 
% \bottomrule
\hline
\label{table:ablation}
\end{tabular}
}
\end{table*}

% \begin{figure*}[t]
%     \centering
%     {\includegraphics[width=0.2\textwidth]{images/478_sgnet.png}} 
%     {\includegraphics[width=0.2\textwidth]{images/478.png}} 
%     {\includegraphics[width=0.2\textwidth]{images/347_sgnet.png}}
%     {\includegraphics[width=0.2\textwidth]{images/347.png}}\\
%     \subfigure[SGNet]{\includegraphics[width=0.2\textwidth]{images/478_sgnet_2d.png}} 
%     \subfigure[CPTNet]{\includegraphics[width=0.2\textwidth]{images/478_2d.png}} 
%     \subfigure[SGNet]{\includegraphics[width=0.2\textwidth]{images/347_sgnet_2d.png}}
%     \subfigure[CPTNet]{\includegraphics[width=0.2\textwidth]{images/347_2d.png}}
%     \caption{Predicted (red) and ground truth (yellow) trajectories for The proposed method and SGNet* on the PIE dataset. Top row shows a sample where the car is moving forward, while the bottom row shows an example where the car is stationary. SGNet* is retrained using the paprameters in~\cite{sgnet}.}
%     \label{fig:qual}
% \end{figure*}

\begin{figure*}[t]
\hspace{-4mm}
    \begin{minipage}{.52\textwidth}
        \centering
        {\includegraphics[width=.44\linewidth]{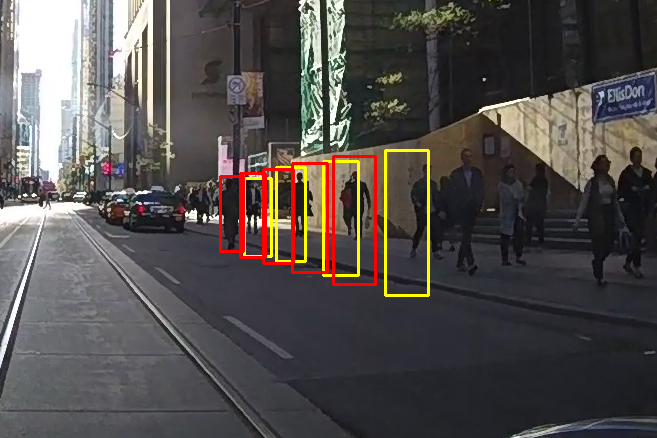}}\hspace{1pt}
        {\includegraphics[width=.44\linewidth]{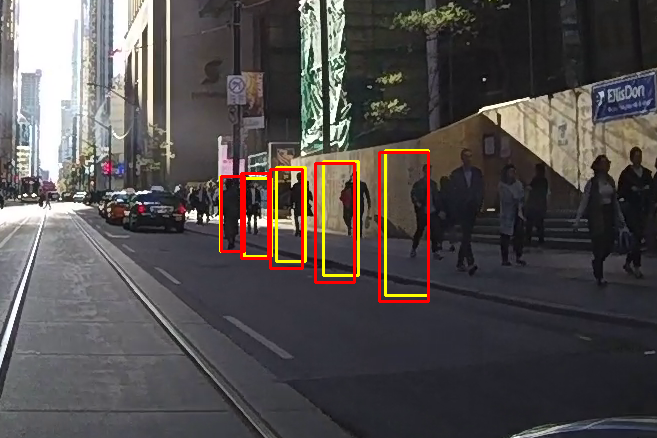}}
        % \caption{First figure with two subfigures}
        \\
        \vspace{2pt}
        \subfloat[SGNet]{\includegraphics[width=.44\linewidth]{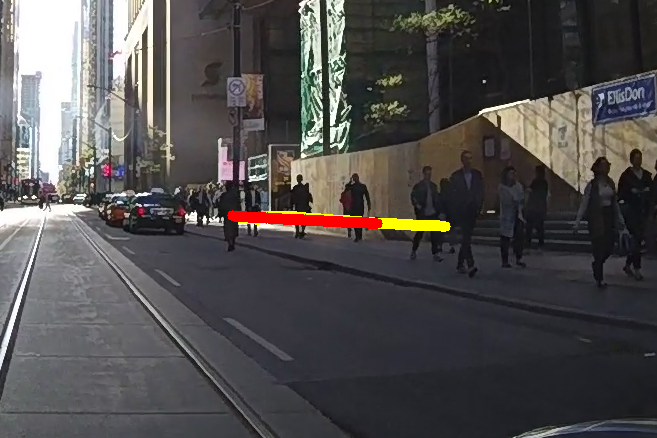}}\hspace{1pt}
        \subfloat[Proposed]{\includegraphics[width=.44\linewidth]{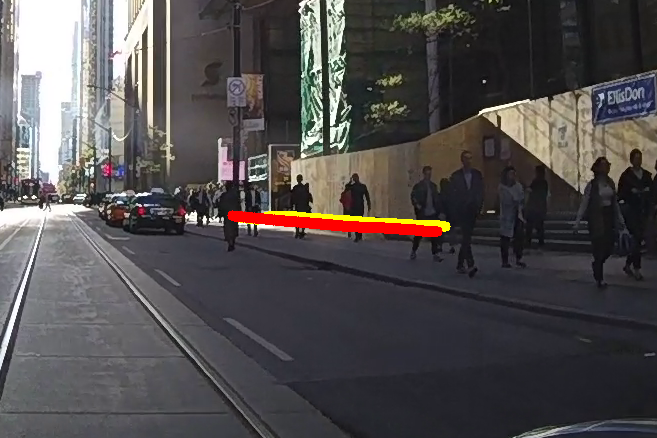}}
        \vspace{2pt}
        \\ (a) Example 1 (trajectory direction $\rightarrow$)
        \label{fig:testa}
    \end{minipage}
    \hspace{-5mm}
    % \hspace{0.05\textwidth}%
    \begin{minipage}{.52\textwidth}
        \centering
        {\includegraphics[width=.44\linewidth]{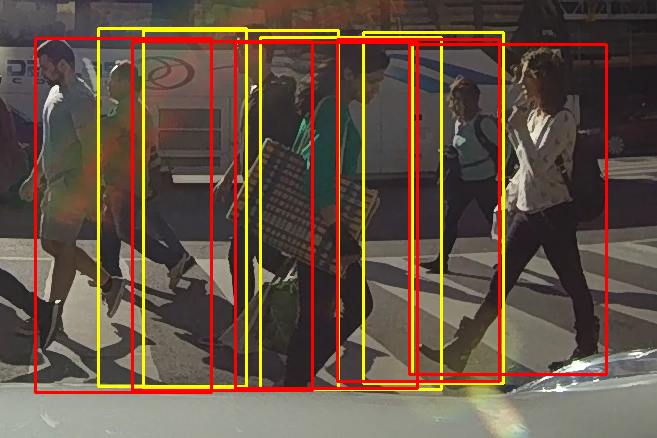}}\hspace{1pt}
        {\includegraphics[width=.44\linewidth]{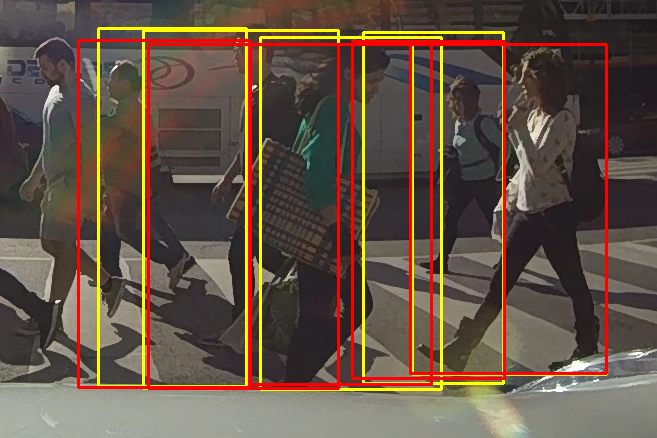}}
        % \caption{First figure with two subfigures}
        \\
        \vspace{2pt}
        \subfloat[SGNet]{\includegraphics[width=.44\linewidth]{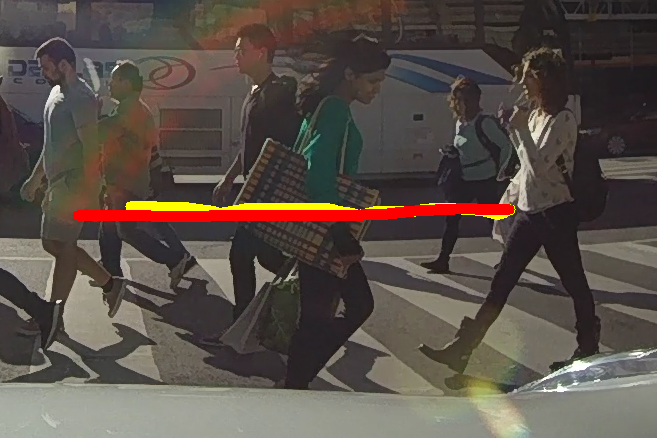}}\hspace{1pt}
        \subfloat[Proposed]{\includegraphics[width=.44\linewidth]{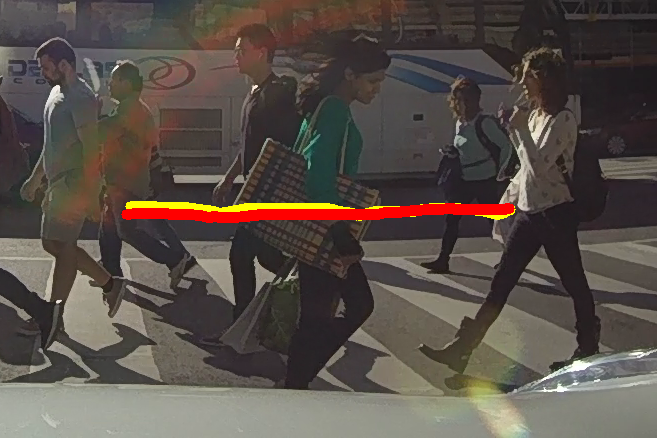}}
        \vspace{2pt}
        \\ (b) Example 2 (trajectory direction $\leftarrow$)
        \label{fig:testa}
    \end{minipage}

    \caption{Predicted (red) and ground truth (yellow) trajectories for the proposed method and SGNet* on the PIE dataset. Example 1 shows a case where the car is moving forward, while example 2 shows an instance where the car is stationary. SGNet is retrained using the hyperparameters in~\cite{sgnet}.}
    \label{fig:qual}
\end{figure*}

\section{Proposed Method}

\noindent\textbf{Problem setup.}
Let the observed location of a pedestrian at time $t$ be denoted by  $\bigl\{ C^x_{t}, C^y_{t}\bigr\}$. We can define a bounding box for the pedestrian, denoted by $X_{t}= \bigl\{ C^x_{t}, C^y_{t}, W_t, H_t\bigr\}$ where $C^x_{t}$ and $C^y_{t}$ are the bounding box center coordinates, and $W_t$ and $H_t$ are the width and height of the box respectively. We aim to develop a model $F$ to predict the future location of the pedestrian, $\widehat{X}_{t+i}= \bigl\{ \widehat{C}^x_{t+i}, \widehat{C}^y_{t+i}, \widehat{W}_{t+i}, \widehat{H}_{t+i}\bigr\}$ for $i \in \{1,\cdots,T_{pred}\}$, based on inputs $X_{t-j}= \bigl\{ C^x_{t-j}, C^y_{t-j},\allowbreak W_{t-j}, H_{t-j}\bigr\}$ for $j \in \{T_{obs}-1,\cdots,0\}$. Here, $t+T_{pred}$ denotes prediction horizon (final instance in the \textit{predicted} trajectory), while $t-T_{obs}+1$ denotes the earliest instance in the \textit{observed} trajectory.

\noindent\textbf{Proposed solution.}
In this work, we aim to use the ego-vehicle speed as contextual information to develop a more effective predictor. This is motivated by previous works~\cite{bayesianlstm, pie} which have shown the speed of the vehicle on which the camera is mounted to be an important factor in determining how the surroundings are perceived. We thus aim to fuse the observed sequence of pedestrian trajectories $\{X_i\}_{t-T_{obs}+1}^{t}$ and corresponding ego-vehicle speeds $\{S_i\}_{t-T_{obs}+1}^{t}$ to predict the future pedestrian locations $\{\widehat{X}_j\}_{t+1}^{t+T_{pred}}$.

To this end, we propose $F$ as a multimodal transformer-based encoder-decoder architecture for pedestrian trajectory prediction from a car's point of view. The architecture for the proposed network is illustrated in Fig.~\ref{fig:proposed}. In our proposed solution, the embeddings for $S_{i}$ and $X_{i}$ are first extracted.
% with a single fully connected layer. 
Next, positional encoding is applied on these extracted features, which are then concatenated together as follows:
\begin{equation}
Z_{X_t,S_t} = P(\psi_1(X_t; \theta_1)) \oplus P(\psi_2(S_t; \theta_2)),
\end{equation}
where $Z_{X_t,S_t}$ are the fused extracted features at timestep $t$, $P(.)$ denotes the positional encoding operator, $\psi_1$ and $\psi_2$ are the feature extractors for location and speed, $\theta_1$ and $\theta_2$ are the trained parameters for the two feature extractors respectively, and $\oplus$ denotes temporal concatenation operation. To further encode these representations and obtain a more abstract embedding with added salience on the important features, we pass the embeddings to a multimodal encoder network consisting of multi-head self-attention, add and normalization layers, and fully connected layers. 

% The proposed decoder architecture is different from the decoder architecture of a standard transformer in two ways. First The proposed decoder does not have a feature extraction module or masked multi-head self-attention module for the decoder input. Further, the input to the proposed decoder during training is a positional encoded empty tensor (i.e. containing all zeros). It may seem counter-intuitive to feed in positional encoded zero values to the decoder, compared to the standard approach of inputting the positional encoded past feature sequences. Nevertheless, we have found this modification to yield superior results, a possible explanation for which can be found in self-supervised learning systems, where learning different pseudo tasks with increased levels of difficulty can help the network learn better representations~\cite{sarkar2020self}. Thus, in certain scenarios, a more challenging task such as we have here, has been shown to yield better results.
Next, in order to generate the predicted trajectory from the learned embedding, we pass the encoded representations to a decoder. A common approach in time-series prediction with transformers is to use some of the information from the ground-truth future values as auxiliary inputs to the decoder during training~\cite{attention, tranforemrtraj}. This strategy allows for a recursive prediction approach during inference, and also aids the final prediction by leveraging past predicted values. In our method, however, we use an empty tensor (i.e., containing all zeros), as the auxiliary input along with the encoded embedding, and achieve better results (see Section~\ref{experiments}). We hypothesize that this strategy allows for the decoder to be trained more effectively by making the task more challenging. This approach also allows for our model to make the predictions for the entire trajectory with a single pass, reducing inference time significantly. Finally, after the decoder, a regressor network is utilized to generate the trajectory from the decoded embedding. Fig.~\ref{fig:proposed} depicts the architecture of our proposed solution. 

\noindent \textbf{Implementation Details.} After feature extraction which consists of a single fully connected layer, the model yields a feature size of 256 for the pedestrian trajectory and 128 for the ego-vehicle speed. In both the encoder and decoder, 16 attention heads are used for the multi-head self-attention modules. Moreover, linear layers of 1024 neurons are used for the fully connected layers. The regressor layer after the decoder is a single fully connected layer. We train our model with a batch size of 128 for 200 epochs, while a learning rate of 0.0005 and an exponential learning rate scheduler is used similar to~\cite{sgnet}. The loss function to train the network is the RMSE between the ground truth and predicted trajectories. All the models were trained and evaluated on an NVIDIA V100 Volta GPU.
% \vspace{-11pt}

\section{Experiments}\label{experiments}

\noindent \textbf{Datasets.} We train our model on two popular datasets, JAAD~\cite{jaad} and PIE~\cite{pie}. They contain 2800 and 1835 pedestrian trajectories respectively, recorded at 30 frames per second. Following the benchmark for these two datasets~\cite{pie, jaad}, they are divided into three sets of 50\%, 10\%, and 40\% for train, validation, and test sets. For both benchmarks, 15 timesteps (0.5s) are observed ($T_{obs}=15$) and 45 timesteps (1.5s) are predicted ($T_{pred}=45$).

\noindent \textbf{Evaluation metrics.} Following~\cite{sgnet, bitrap, pie}, we use the Mean Squared Error (MSE) between the two opposing corners of the ground truth and predicted bounding boxes, as well as their centers (CMSE), to evaluate the proposed model. The MSE and CMSE between the predicted and ground truth trajectory timesteps (CFMSE) are also calculated to determine the accuracy of the predictions at the end of the trajectories. Whereas MSE and CMSE average the errors across all frames in the trajectory, CFMSE only considers the final frame.
% \noindent \textbf{Benchmarks.}
% In the following section, we will compare our solution against prior state-of-the-art techniques, namely ???, ???, ???

\noindent \textbf{Results.}
\label{sec:pagestyle}
The evaluation results for the proposed method as well as the prior state-of-the-art are presented in Table~\ref{table:eval_res}. We compare our solution against prior techniques~\cite{bayesianlstm,folx,pie,bitrap, sgnet}. Moreover, in order to visualize and qualitatively compare the results with the latest state-of-the-art (SGNet), we re-trained the model and denote its results with SGNet*. Finally, we implement a standard transformer based on~\cite{attention} for further comparison. We observe that our proposed method outperforms all prior works on both datasets by considerable margins. We also observe that while the superior performance of our method is consistent across all  prediction windows (0.5, 1.0, and 1.5 seconds), our method shows more percentage improvement for longer prediction windows. This is likely due to pedestrian parallax caused by ego-vehicle speed, which becomes more significant over longer prediction windows.

\noindent \textbf{Ablation.} 
To evaluate the impact of the two modalities used in our solution, we perform ablation studies, and present the results in Table~\ref{table:ablation}. First, we remove the ego-vehicle speed from our model. Here we observe a drop in performance across all the experiment setups, with the exception of the MSE metric for short-term windows (0.5 seconds in both datasets and 1.0 seconds in the case of JAAD). This observation was to be expected since JAAD has categorical speed labels as opposed to Pie, which includes numerical values for speed, making the effect of speed less significant. The overall degradation of the network after eliminating speed indicates the impact of our approach for longer-term predictions. We also observe that the performance of the network without the help of speed is better compared to prior works, which indicates the superior performance of the proposed single-pass transformer-based architecture. Next, we ablate the location and use only speed for prediction. As expected, a significant drop is observed as the model is not able to make valid predictions without any trajectory observations. 

\balance

A qualitative analysis is provided in Fig.~\ref{fig:qual}. We provide two examples, where Example 1 depicts a case where the vehicle is moving, while Example 2 is obtained from a stationary vehicle. In this Fig. We compare our results against the retrained SGNet*, where we observe that our method achieves more accurate results in terms of both bounding boxes (top row) and their centers (bottom row).

Table~\ref{table:hyp} presents the inference time (in milliseconds) for $T_{pred}=45$ predicted samples based on $T_{obs}=15$ observations, as well as the number of trainable parameters in the proposed approach, SGNet~\cite{sgnet}, BiTrap~\cite{bitrap}, and PIETraj~\cite{pie}. The computation times for all methods were evaluated on a local NVIDIA GTX 3090 GPU. We observe that our proposed method is 
% both smaller and 
significantly faster than the previous methods, making it more suitable for real-time mobile and vehicular edge deployment.

\begin{table}[!t]
\centering
\footnotesize
\caption{Comparison between the number of parameters, and inference time of our method and retrained SGNet* on PIE.}
\label{table:hyp}
\resizebox{0.85\columnwidth}{!}{
\begin{tabular}{c c c} 
\hline
Method & Num. of parameters & Inference time (ms)\\
% \hline
\hline
  PIETraj~\cite{pie} &  17.7M & 374\\ %change citations
  BiTraP~\cite{bitrap} & 1.5M & 16\\
%  SGNet~\cite{sgnet} & 7622406 & 515\\
  SGNet~\cite{sgnet} & 4.3M & 470\\
%  \hline
%  CPTNet-single-loop & 6994948 & 4\\ 
Proposed & 2.9M & 2\\
% CPTNet-45 & 3.4M & 34\\ 
\hline
\end{tabular}
}
\end{table}

% \begin{figure*}[!tbp]
%   \centering
%   \includegraphics[width=0.4\textwidth]{images/qual.pdf}
%   \caption{visualizations}
%   \label{fig:proposed}
% \end{figure*}

% \begin{table*}[!tbp]
% \centering
% \caption{Ablation studies on PIE and JAAD dataset}
% \begin{tabular}{cccc|ccccc|cccccc}
% % \toprule[1pt]\midrule[0.3pt]
% \hline
% \multicolumn{4}{c|}{Method}   & \multicolumn{5}{c|s}{\textbf{PIE}} & 
% \multicolumn{5}{c}{\textbf{JAAD}}  \\
% \cmidrule(r){1-4}\cmidrule(r){5-9} \cmidrule(r){10-14}
% Loc & Joint & Loc & Sp & \multicolumn{3}{c}{MSE} &  CMSE & CFMSE &
% \multicolumn{3}{c}{MSE} &  CMSE & CFMSE  \\
%  \cmidrule(r){5-7} \cmidrule(r){10-12}
%  Enc & Enc & Dec & Dec & 0.5s & 1s & 1.5s & 1.5s & 1.5s & 0.5s & 1s & 1.5s & 1.5s & 1.5s\\
% \midrule
%  \cmark & \xmark &  \cmark & \xmark & 33 & 132 & 436 & 408 & 1725 & 75 & 320 & 1044 & 998 & 4084   \\ 
%  \cmark &  \cmark &  \cmark & \cmark & - & - & - & - & - & - & - & - & - & - \\ 
%  \cmark &  \cmark & \cmark &  \cmark & \textbf{34} & \textbf{128} & \textbf{400} & \textbf{373} & \textbf{1,548} & \textbf{78} & \textbf{313} & \textbf{985} & \textbf{936} & \textbf{3777}\\ 
% % \bottomrule
% \hline
% \end{tabular}
% \end{table*}

\section{Conclusion}
\label{sec:typestyle}
In this paper, we proposed a multimodal encoder-decoder transformer network that learns to predict pedestrian trajectories based on observed trajectory and ego-vehicle speed data. Keeping the goal of edge deployment in mind, our model is designed to make the prediction in a single pass and avoid single-step prediction approach. We experiment our method on two popular public datasets, PIE and JAAD, and observe superior results across both datasets and various prediction time horizons. Moreover, ablation studies demonstrate the important impact of using the ego-vehicle speed in our solution. Lastly, we compare the parameters and inference time of our method against prior state-of-the-art and observe significant improvements for both factors. 
% The extracted features from two modalities of vehicle speed and trajectory are concatenated, and fed to the joint encoder. Evaluations and ablation studies confirm that using vehicle speed in predicting future trajectory can have a positive effect. Further, the proposed modified decoder architecture is capable of producing the future rejectories in one step, without having to wait for the previous timestep to be predicted. This causes the network to be significantly faster than the state-of-the-art model on the two PIE and JAAD datasets. The evaluation results as well as the visualizations confirm the superiority of the proposed method against the state of the art.

\vspace{2mm}
\noindent \textbf{Acknowledgement.}
Thanks to Geotab Inc., the City of Kingston, and NSERC for their support of this work. We also thank Compute Canada for the GPU resources.
\vfill\pagebreak

% References should be produced using the bibtex program from suitable
% BiBTeX files (here: strings, refs, manuals). The IEEEbib.bst bibliography
% style file from IEEE produces unsorted bibliography list.
% -------------------------------------------------------------------------
\bibliographystyle{IEEEbib}
\bibliography{main}

\end{document}